\documentclass{arxiv}

\usepackage{makeidx}         
\usepackage{multicol}        
\usepackage{multirow}

\usepackage{graphicx}
\usepackage{epstopdf}
\usepackage{subfigure}
\usepackage{amssymb}
\usepackage{color}
\usepackage{epsfig} 
\usepackage{amsmath}
\usepackage{balance}
\usepackage{url}

\usepackage{url}

\usepackage[ruled]{algorithm2e}

\usepackage{amsmath} 

\newcommand{\comment}[1]{}

\usepackage{pdfpages}

\title{Deep Leaf Segmentation Using Synthetic Data}

\addauthor{Daniel Ward}{Daniel.Ward@data61.csiro.au}{1}
\addauthor{Peyman Moghadam}{Peyman.Moghadam@data61.csiro.au}{2}
\addauthor{Nicolas Hudson}{Nicolas.Hudson@data61.csiro.au}{3}

\addinstitution{
 Robotics and Autonomous Systems \\
The Commonwealth Scientific and Industrial Research Organisation (CSIRO), Data61\\
 Brisbane, Australia
}
\runninghead{Ward, Moghadam, Hudson}{Deep Leaf Segmentation Using Synthetic Data}

\def\etal{\emph{et~al}\bmvaOneDot}

\begin{document}

\maketitle 

\begin{abstract}
Automated segmentation of individual leaves of a plant in an image is a prerequisite to measure more complex phenotypic traits in high-throughput phenotyping. Applying state-of-the-art machine learning approaches to tackle leaf instance segmentation requires a large amount of manually annotated training data. Currently, 
the benchmark datasets for leaf segmentation contain only a few hundred labeled training images. In this paper, we propose a framework for leaf instance segmentation by augmenting real plant datasets with generated synthetic images of plants inspired by domain randomisation. We train a state-of-the-art deep learning segmentation architecture (Mask-RCNN) with a combination of real and synthetic images of \textit{Arabidopsis} plants. Our proposed approach achieves 90\% leaf segmentation score on the A1 test set outperforming the-state-of-the-art approaches for the CVPPP Leaf Segmentation Challenge (LSC). Our approach also achieves 81\% mean performance over all five test datasets. 

\end{abstract}

\section{Introduction}
\label{sec:intro}

%
%

To achieve sustainable agriculture, we need to expedite the breading of new plant varieties which consume less water, land or fertilizer and have greater resistance to parasites and diseases while producing greater crop yields. 
Plant phenotyping is the process of relating plant varieties to genotype and environmental conditions and how these affect the observable plant traits.

In recent years, several computer vision and machine learning techniques \cite{scharr2016leaf, moghadam2017plant} have been proposed to increase the throughput of non-destructive phenotyping. 
Automatic, non-destructive extraction of plant parameters such as plant height, shape, leaf area index (LAI) or growth rate throughout the crop growth cycle are essential for rapid phenotype discovery and analysis. 
Phenotypic measurements are often done at two levels: plant-level measurements (projected LAI, height, plant architecture) or leaf-level measurements (individual leaf area, leaf count, leaf growth rate). 
This paper focuses on the former category.   

Despite substantial progress, segmentation of individual leaves (leaf instance segmentation) remains extremely challenging owing to the variability in leaf shapes and appearance over the life-cycle of the deformable object (plant). 
Another major challenge is that leaves partially overlap and occlude each other as new leaves grow.

Currently, leading instance segmentation techniques \cite{he17} based on deep convolutional neural networks require huge amounts of annotated training data. 
Manual annotation of images for a task such as instance segmentation is often time consuming and expensive. 
For leaf instance segmentation there are only a few annotated datasets available and the size of these datasets is very small (a few hundred images)~\cite{minervini2016finely}.
In this paper we use the CVPPP dataset from the leading Leaf Segmentation Challenge (LSC). It is a small dataset of top-down 2D visible light images of \textit{Arabidopsis} and tobacco plants from an indoor high throughput plant phenotyping system. 
This dataset only contains 27 images of tobacco and 783 \textit{Arabidopsis} images with pixel-level leaf segmentation labels.

\begin{figure}[t]
  \centering
    \subfigure[]{
        \includegraphics[width=0.22\textwidth]{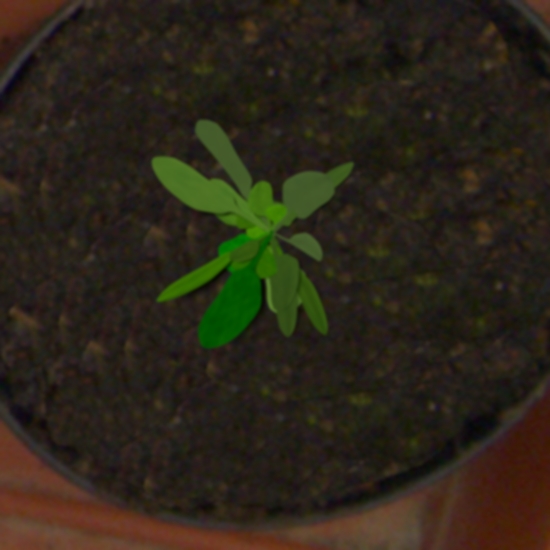}
    }
  	\subfigure[]{
        \includegraphics[width=0.22\textwidth]{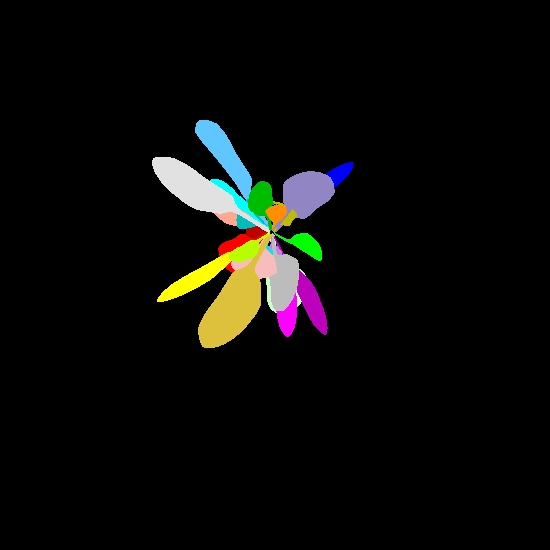}
    }
    \subfigure[]{
        \includegraphics[width=0.22\textwidth]{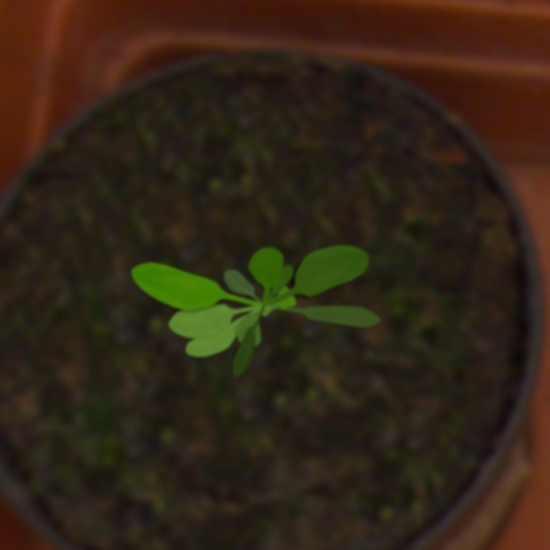}
    }
  	\subfigure[]{
        \includegraphics[width=0.22\textwidth]{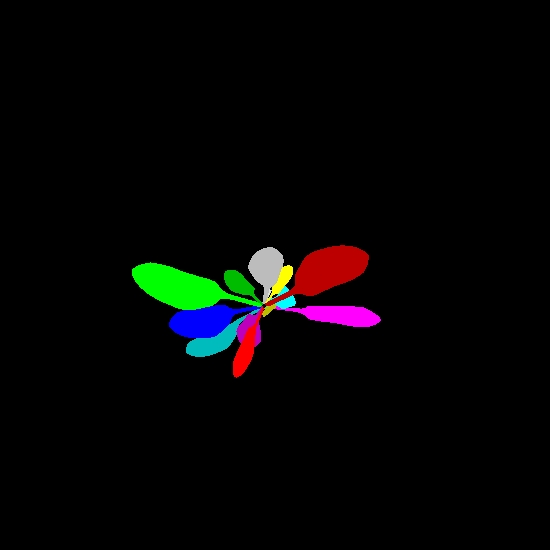}
    }
  \caption{Example top down view of synthetic \textit{Arabidopsis} plants (a, c) and their segmentation labels (b, d).}
  \label{fig:exampleSynthData}
\end{figure}

To address these problems, we propose a framework to augment limited annotated training data of real plants with synthetic plant images inspired by domain randomisation \cite{tremblay2018}. Synthetic plant images come without the cost of data collection and annotation (examples shown in Figure \ref{fig:exampleSynthData}). 
Our main contributions can be summarised as follows:
\begin{itemize}
\item We outperform the state-of-the-art leaf segmentation approach by augmenting real data with synthetic data.
\vspace{-2mm}
\item This is the first time synthetic data has been used for the CVPPP Leaf Segmentation Challenge using a novel plant modeling system.
\vspace{-2mm}
\item We investigate the application of domain randomisation concepts on deformable objects (plants) and found that sampling realistic foreground and background textures improved results.
\end{itemize}

\section{Related Work}
\label{sec:related_work}

There has been significant progress in the vision community on object instance segmentation algorithms, which requires the detection of each object (or leaf) in the image, and then the precise segmentation of each instance. 

In this paper, we consider the Mask-RCNN \cite{he17} framework, as it achieved state of the art single-model results on the COCO
instance segmentation task \cite{lin2014} in 2017, and feature pyramid derivatives of it remain the competition leaders. With the addition of the branch for predicting segmentation masks, Mask-RCNN is a low overhead extension of Faster-RCNN, which recently also demonstrated the best detection
performance among a family of object detectors \cite{huang17}.

Training detection or segmentation algorithms takes vast amounts of labeled data: COCO instance segmentation training has 123,287 images; ImageNet \cite{russakovsky2015} detection has over one million images with annotated object bounding boxes, although some classes in it, such as \emph{strawberry}, will appear in only hundreds of those images. One of the few large scale data sets where objects are broken down further into segmented components is ADE20K \cite{zhou2017scene}, with 20,210 images with all objects, and components within segmented.  

In comparison, the CVPPP LSC uses 810 images for training, each with a single plant with multiple leaf instances. Increasing the size of this training data set would lead to increased test set performance. Recent work in classifying birds \cite{vanhorn2017}, while varying the training data set from 10 to 10,000 images, suggested a rule where the test set error drops by a factor of 2x each for every 10x increase in the number of training images. 

Given a limited training dataset, where it is hard or expensive to annotate more real world data, recent work in computer vision looks at using simulated data to train networks. It known that the \emph{reality gap}, or the inability of renders to create the statistics of the real world, lead to results where naive training and evaluations on synthetic images do not transfer to the real world. 
The literature has several approaches to incorporating synthetic imagery including: Mixing generated objects over real world scenes \cite{Dwibedi2017,alhaija2017}; domain randomisation \cite{tremblay2018}, where parameters of the simulator creating the images are varied in non-realistic ways to make the networks invariant to them; creating photo-realistic renders \cite{alhaija2017}, where substantial effort is made to reduce the reality gap directly; and General Adversarial Networks (GANs) \cite{Bousmalis2016} to automatically learn and compensate for important differences between simulated and real world data. Freezing a pre-trained feature extractor layer in the networks and using simulation to only train output layers \cite{hinterstoisser2017} has shown amazing results for a limited set of rigid objects, when in fact the pre-trained features are expressive enough. Hinterstoisser~\etal~\cite{hinterstoisser2017} found that this resulted in features closer to those extracted from real images being extracted from the synthetic images.

To the best of our knowledge, synthetic data has been used twice before in conjunction with the CVPPP datasets. In the absence of generating synthetic segmentation labels, both approaches focused on estimating the count of leaves in each image. Furthermore, each method's synthetic data was rendered without a background. 
ARIGAN~\cite{giuffrida2017} presents a data driven approach where a GAN learns the distribution of the CVPPP A1 subset dataset and generates new samples to an expected leaf count. GANs are renowned for more accurately modeling texture than geometry and this is evident in the ARIGAN synthetic \textit{Arabidopsis} dataset.
In the second synthetic data approach, Ubbens \etal \cite{ubbens2018use} leverages a model based on the L-system formalism. L-systems are commonly used to describe the structure and development of plants. The design of their \textit{Arabidopsis} L-system model involved defining plant attributes including leaf shape and inclination angle. 
These attributes were defined using functions drawn by manually placing control points within an L-systems library~\cite{prusinkiewicz2002art}.

%
%
\section{Method}
\label{sec:syntheticData}
This work augments and compares performance of real world training to using synthetic leaf images. Leveraging ideas in the domain randomization literature, leaf textures are randomised and simulated data is overlayed onto random real world scenes. In addition, we evaluate freezing feature layers in models like in \cite{hinterstoisser2017}. The work here extends these ideas to deformable plant models where each object is made up of many leaf components, and does not adhere to the rigid body models used in most synthetic data approaches.

\subsection{Synthetic Data Generation for Arabidopsis}
\label{subsec:synthData}
%
%

%
%
%
\begin{figure}
	\centering
	\includegraphics[width=12cm]{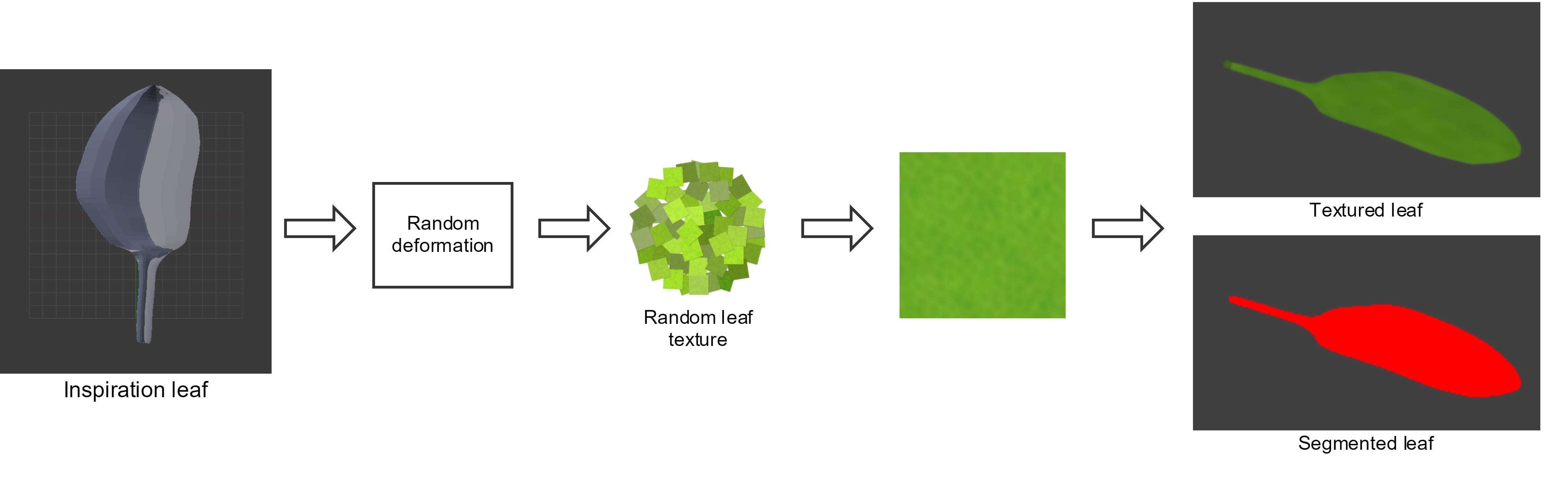}
    \vspace{-3mm}
    \caption{The synthetic leaf generation pipeline. Each leaf in a synthesised plant is randomly deformed, textured and positioned.}
    \label{fig:leafGeneration}
    \vspace{-5mm}
\end{figure}

\begin{figure}[!htpb]
	\centering
	\includegraphics[width=11cm]{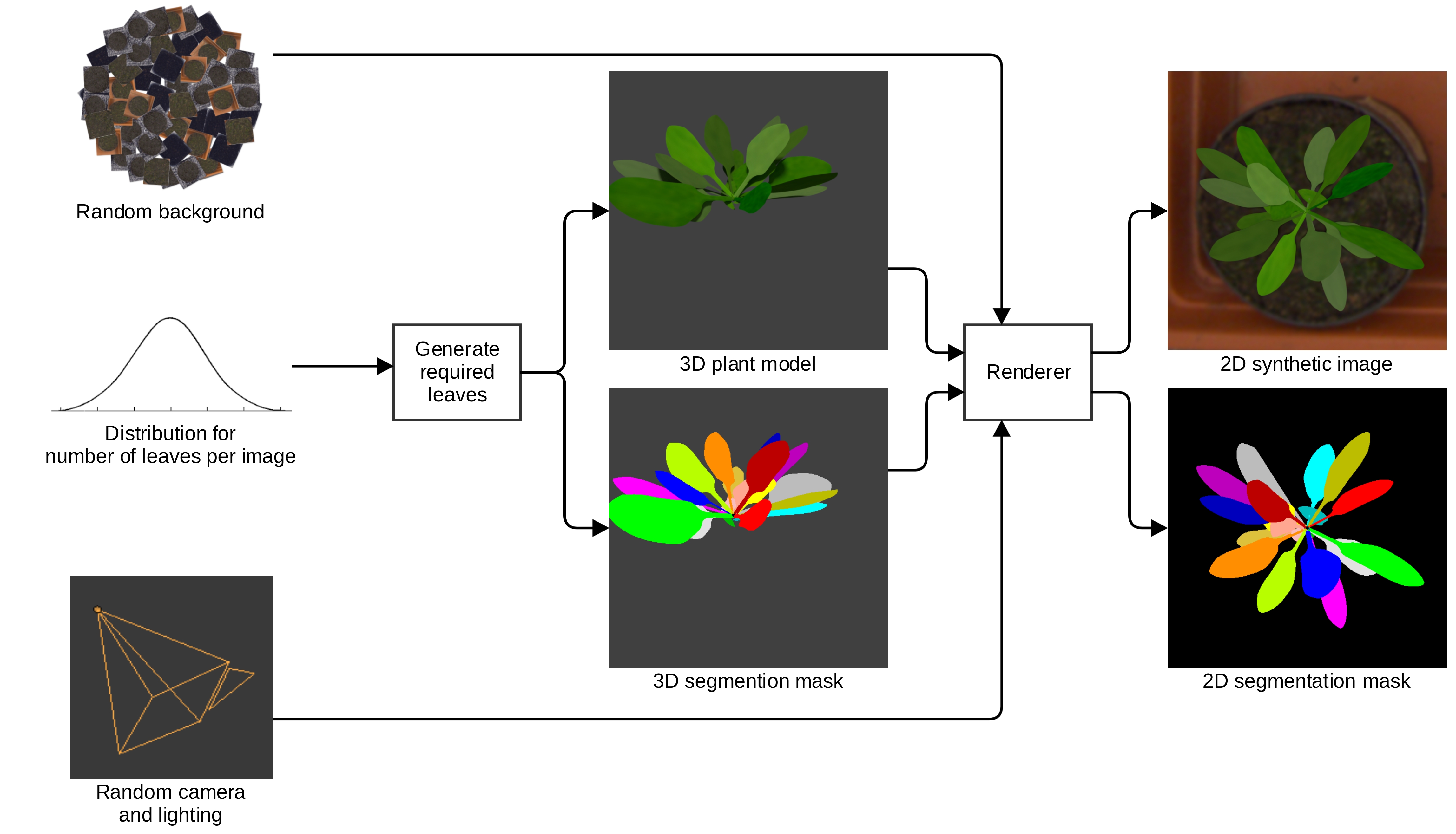}
    \caption{The synthetic data pipeline. A 3D plant model of synthetic leaves (Figure \ref{fig:leafGeneration}) is assembled and rendered over random plant pot background.}
    \label{fig:plantGeneration}
    \vspace{-5mm}
\end{figure}

Our synthetic \textit{Arabidopsis} generation pipeline (Figures \ref{fig:leafGeneration} and \ref{fig:plantGeneration}) begins with manual specification of geometric properties. This is done by defining an \textit{inspiration leaf} rather than multiple functions like in L-systems~\cite{ubbens2018use}. First, an inspiration leaf was designed by tracing a randomly chosen \textit{Arabidopsis} leaf image in Blender to produce a 3D mesh by adding and adjusting manipulation points. This was done only once for all synthetic data used in this study. Figure \ref{fig:leafGeneration}, displays the steps involved in generating a single leaf from the original inspiration.

In order to model leaves of different shape and size, every leaf is randomly scaled along each axis independently. This process differs from other applications such as synthetic cars, where the resulting significant changes to object aspect ratio would be undesirable.
To provide variation in texture between leaves, each one is rendered with a different texture. Thirty leaf textures were extracted from the CVPPP training datasets and augmented using the data augmentation as described in Section \ref{subsec:training} with the addition of random adjustments to image exposure.

Being rosettes, the leaves of an \textit{Arabidopsis} plant are arranged circularly. Furthermore, all leaves are at a similar height and remain close together \cite{mundermann2005quantitative}. Following this, our model can be intuitively thought of as an arrangement of leaves stemming out from a sphere. The location of leaf stems is most likely at the equator and equally likely at any point around the equator in the $xy$ plane.  With the leaf initialised at the origin in the $xy$ plane, the final position of the leaf is defined by rotating the mesh around the $x$, $y$ and $z$ axes independently. To produce the rosette, the rotation around the $z$ axis is sampled from a uniform distribution, $U(0, 2\pi)$. Similarly, the leaf pitch and roll are sampled from $U(\frac{-\pi}{4}, \frac{\pi}{4})$ and $U(0, \frac{\pi}{4})$ respectively to subtly randomise leaf pose. Uniform distributions were used to provide a wide variety of leaf positions. This process is shown in Figure \ref{fig:plantGeneration} and repeated multiple times according to a normal distribution of leaves per plant centered around the mean leaf count in the CVPPP A1 subset dataset but with wider variance (Figure \ref{fig:leafCounts}).
%
%
%
Figure \ref{fig:plantGeneration} defines the plant generation pipeline. The number of leaves in a generated plant is first sampled from a provided distribution and the leaf generation pipeline (Figure \ref{fig:leafGeneration}) is called. A top down view of the 3D plant model is then rendered to simulate the CVPPP data. Like \cite{tremblay2018}, the camera and lighting positions were varied however only a single light source was used. The camera position randomisations were restricted to ensure a top down view was still obtained.
Using a sample from each CVPPP dataset, A1-A4, we extracted the background images to obtain 4 plant pot background samples. These were then processed using the data augmentation described in Section \ref{subsec:training} and used for the render backgrounds. Following \cite{hinterstoisser2017}, we applied Gaussian blurring to the rendered plant to better integrate it into the real world scene (background).

Our synthetic data pipeline was designed in Blender (v2.79b). We use the Cycles renderer to produce the synthetic images with Lambertian and Oren-Nayar diffuse reflection shading. The segmentation labels were rendered using the native Blender renderer with all shading and anti-aliasing disabled to ensure their integrity. 

\subsection{Training \& Evaluation}
\label{subsec:training}

We used the Matterport\footnote{\url{https://github.com/matterport/Mask\_RCNN}} Mask-RCNN implementation for our experiments~\cite{he17}. Mask-RCNN is a two-stage model, with a feature extractor feeding into a Region Proposal Network (RPN) and then into three heads producing box classification, box regression, and an object mask. We use a ResNet101 backbone with a Feature Pyramid Network for Mask-RCNN. Unlike the original implementation, we use 256 (opposed to 512) regions of interest per image during training.

In each experiment, the model was initialised with weights trained on the COCO dataset. It was then retrained on a combination of real world and synthetic data. Data augmentation is a common practice to improve model robustness and prevent over fitting. In all experiments random: left-right flipping; top-bottom flipping; rotation; zooming and cropping were applied to the training images. Data augmentation was applied to real and synthetic datasets. Our training procedure consisted of splitting the data into 80\%, 20\% training and cross validation sets respectively; batch normalisation across a batch size of 6 and leveraging early stopping. In experiments where both real and synthetic data were used for training, each batch was balanced such that 50\% of images came from each dataset respectively.

To evaluate our methods, we compete in the CVPPP Leaf Segmentation Challenge (LSC)\footnote{\url{https://competitions.codalab.org/competitions/18405}}. It consists of three \textit{Arabidopsis} plant (A1, A2, A4) datasets and a dataset of young tobacco plant images, A3.  Training sets for A1, A2, A3, A4 contain 128, 31, 27, 624 images while the test sets have 33, 9, 65, 168 images respectively. Additionally, test set, A5, is also provided which combines images from all testing datasets in order to evaluate the generalisation of proposed machine learning techniques. 
In order to compare our leaf instance segmentation performance with the current state-of-the-art approaches from the leaf segmentation Challenge (LSC), we only use the A1 dataset for training.

Using the method described in Section \ref{subsec:synthData}, three synthetic Arabidopsis datasets were generated. Each contained 10,000 labeled images and are defined as follows. 
In \textit{Synthetic-plant}; all leaves in all plants contain the same green texture.
In \textit{Synthetic-leaf}; Each leaf in a plant is rendered with a random different green texture as shown in Figure \ref{fig:leafGeneration}.
Finally, \textit{Synthetic-COCO} is generated following the same process however the leaf textures are sampled from the COCO dataset. 
\textit{CVPPP-A1} was the real dataset used during the experiments which consists of the 128 images from the CVPPP A1 dataset. 
The training procedure matching real and synthetic images per batch described above was named \textit{Synthetic+real}.

\section{Results \& Discussion}
\label{sec:results}

In this section we present our performance on the CVPPP LSC, compare our results to the state-of-the-art and investigate the limitations of the synthetic datasets.
Table~\ref{tab:cvpppResults} displays the results for each experiment and dataset in Section  \ref{subsec:training}. The results shown are performances on the 5 test sets of the CVPPP LSC. Segmentation performances are quoted in the competition metric, symmetric best dice (SBD) score. Note, the column titled \textit{Mean} presents the mean per test image performance across all test sets as provided by the competition and not the mean across the table row.
%
%
\begin{table}[!htpb]
\begin{center}
\begin{tabular}{|l|c|c|c|c|c|c|}
\hline
\multirow{2}{*}{Training Regime} & \multicolumn{6}{|c|}{CVPPP test set (SBD)}\\
\cline{2-7}
 & A1 & A2 & A3 & A4 & A5 & Mean \\
\hline
CVPPP-A1 (real) & 87 & 71 & \textbf{59} & 73 & 70 & 71 \\ 
Synthetic-plant & 81 & 69 & 41 & 84 & 73 & 74 \\ 
Synthetic-leaf & 82 & 74 & 32 & 86 & 75 & 74 \\ 
Synthetic-COCO & 57 & 48 & 25 & 48 & 43 & 44 \\ 
Synthetic+real & \textbf{90} & \textbf{81} & 51 & \textbf{88} & \textbf{82} & \textbf{81} \\
\hline
\end{tabular}
\end{center}
\caption{CVPPP LSC results. Mask-RCNN trained on real data (CVPPP-A1), synthetic data (-plant, -leaf and -COCO) and both real and synthetic (Synthetic+real) data. Segmentation scores are quoted in symmetric best dice (SBD).}
\label{tab:cvpppResults}
\end{table}

We achieve the best results on A1, A2, A4 and A5 when training on both synthetic and real data. Best performance on the A3 subset, however, is obtained by training on real data only. The generalisation of \textit{Synthetic+real} is shown by a significantly higher mean result. Furthermore, the effect of varied texture can be seen in \textit{Synthetic-plant}, \textit{Synthetic-leaf} and \textit{Synthetic-COCO}, where training on randomly textured leaves (\textit{Synthetic-COCO}) achieved the lowest performance. A performance increase is seen when training on plants with different green textured leaves (\textit{Synthetic-leaf}) on all test sets except A3 where \textit{Synthetic-plant} excels.

%
%
\begin{figure}[!htpb]
  \centering
    \subfigure[RGB]{
        \includegraphics[width=0.2\textwidth]{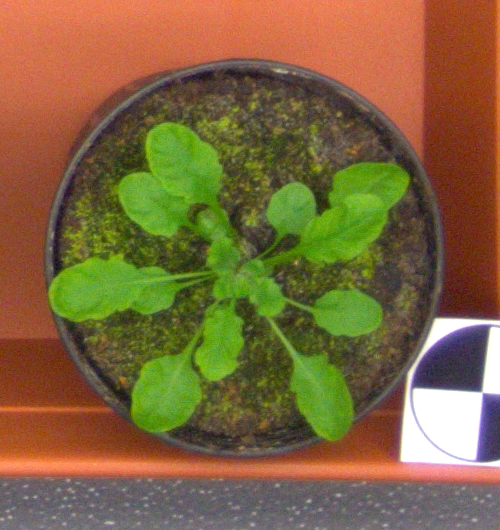}
    }
    \hspace{-1mm}
  	\subfigure[Synthetic+Real]{
        \includegraphics[width=0.2\textwidth]{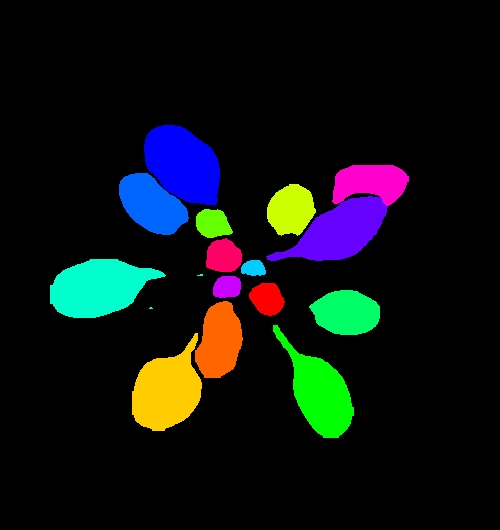}
    }
    \hspace{-1mm}
    \subfigure[CVPPP-A1]{
        \includegraphics[width=0.2\textwidth]{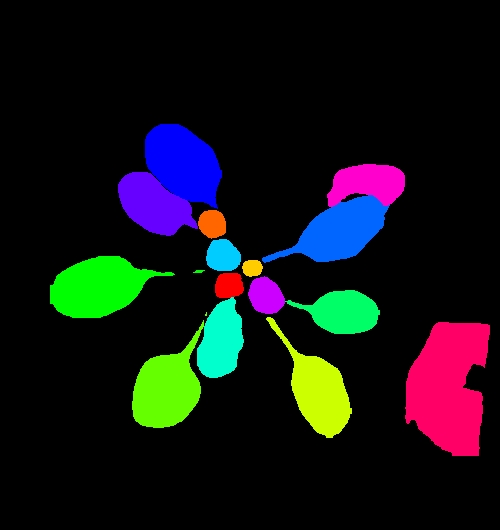}
    }
    \hspace{-1mm}
  	\subfigure[Synthetic-leaf]{
        \includegraphics[width=0.2\textwidth]{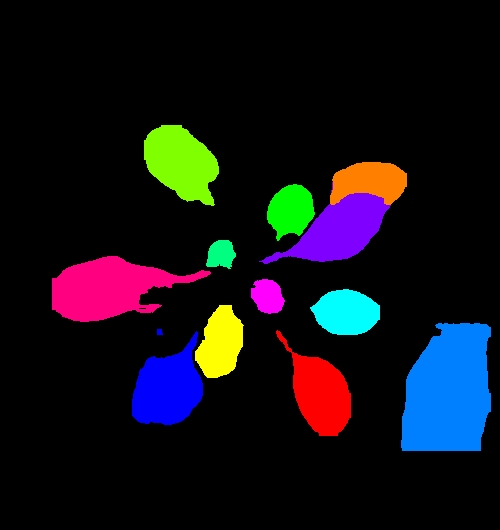}
    }
    \caption{Segmentations of plant 117 (a) from the A1 test set. Predictions are presented from models trained on both synthetic and real data (b) and only real data (c) or synthetic data (d). Only the model trained on both synthetic and real data (b) avoided segmenting the background artifact (fiducial marker). Note that no fiducial markers appear in the real CVPPP A1 or any of the synthetic training datasets.}
    \label{fig:cvpppA1_plant117}
    \vspace{-1mm}
\end{figure}

Figure \ref{fig:cvpppA1_plant117} compares the CVPPP A1 test set prediction for plant 117 from the A1 test set. The artifact in the background of the image, has been incorrectly segmented by both models trained on real only and synthetic data only. The model trained on both real and synthetic data, however, detects more of leaves and does not incorrectly segment the background artifact. No fiducial markers, the artifact, appear in any training data, real or synthetic. The false positive segmentation is likely explained by the curved geometry and edges of the object. Further, a possible reason for the model correctly not segmenting the artifact is the greater variance in geometry in the \textit{Synthetic+real} dataset.

In addition to the results in Table \ref{tab:cvpppResults}, we experimented with freezing the feature extractor layers of Mask-RCNN when training on synthetic data as in \cite{hinterstoisser2017} (see Section \ref{sec:related_work}). This lead to degraded results, similar to \cite{tremblay2018}. We are not confident if this is because of using flexible and deformable plant models as opposed to rigid CAD models that match the test set, or if there is some deficiency in our methodology, as we were constrained in producing synthetic sets at the scales used in \cite{hinterstoisser2017}.

%
%

\begin{table}[!htpb]
\begin{center}
\begin{tabular}{|l|c|c|c|}
\hline
\multirow{2}{*}{Method} & \multicolumn{3}{|c|}{CVPPP test set (SBD)}\\
\cline{2-4}
 & A1 & A2 & A3 \\ 
\hline
RIS + CRF \cite{romera2016recurrent} & 66.6 & - & - \\ 
MSU \cite{scharr2016leaf} & 66.7 & 66.6 & \textbf{59.2} \\ 
Nottingham \cite{scharr2016leaf} & 68.3 & 71.3 & 51.6 \\
Wageningen \cite{yin2014multi} & 71.1 & 75.7 & 57.6 \\
IPK \cite{pape20143} & 74.4 & 76.9 & 53.3 \\
Salvador \etal \cite{cSalvadore} & 74.7 & - & - \\
Brabandere \etal \cite{de2017semantic} & 84.2 & - & - \\
Ren \etal \cite{ren2017end} & 84.9 & - & - \\
Ours & \textbf{90.0} & \textbf{81.0} & 51.0 \\
\hline
\end{tabular}
\end{center}
\caption{CVPPP LSC results. We compare the performance of our best performing model (\textit{Synthetic+real}(A1)) to existing approaches and outperform the A1 and A2 state-of-the-art. We achieve comparable results for A3 (\textit{Tobacco}). Note the real data used to train our model came solely from A1 (\textit{Arabidopsis}). To the best of our knowledge results for A4 and A5 have not yet been published and are hence omitted. Segmentation scores are quoted in symmetric best dice (SBD).}
\label{tab:a1Results}
\end{table}

%
%

Our proposed approach (trained on real and synthetic data) achieves 90\% SBD score outperforming the state-of-the-art leaf instance segmentation algorithms (See Table \ref{tab:a1Results}) for the A1 test set. The real data used to train our model came solely from the A1 data set. Our model also achieves state-of-the-art and comparable performance on A2 and A3 respectively. The A3 dataset contains images of tobacco plants rather than \textit{Arabidopsis} and unlike the compared methods, our model is only trained on images of \textit{Arabidopsis} plants. The reduced performance is explored below.
The networks trained only on synthetic data, on average outperformed the models using only the limited A1 training data set (128 images). This likely due, in a large part, to the increase in variance across the distribution size and position of leaves in the synthetic set, and the much larger size of the data set itself (10,000 images vs 128). The positional variation in the synthetic dataset is also likely why the synthetically trained models perform particularly well on the A4 test set, which has larger range of plant age ranges than A1. These reasons also explain the incorrect segmentation phenomena shown in Figure \ref{fig:cvpppA1_plant117}.

%
%

\begin{figure}[!htpb]
\vspace{-2mm}
  \centering
    \subfigure[Real Data\hspace*{5.7em}]{
        \includegraphics[width=0.3\textwidth]{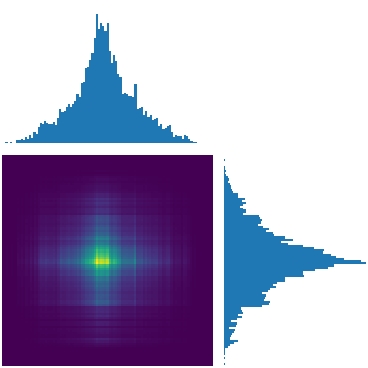}
    }
    \hspace{0.1\textwidth}
  	\subfigure[Synthetic Data\hspace*{5.7em}]{
        \includegraphics[width=0.3\textwidth]{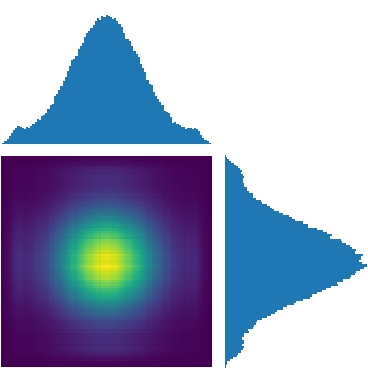}
    }
    \vspace*{2mm}
     \caption{The distribution of leaf centroid locations in the training images. A wider distribution of leaf locations is presented in the synthetic data (b) while preserving the center bias. Note a uniform distribution across the image could be achieved by tuning the synthetic data pipeline.}
    \label{fig:leafLocDist}
    \vspace{-2mm}
\end{figure}

\begin{figure}[htpb]
  \centering
    \subfigure[Real Data]{
        \includegraphics[width=0.4\textwidth]{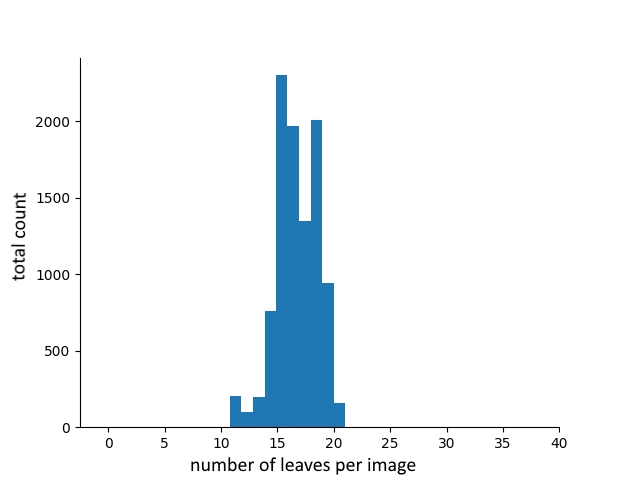}
    }
    \hspace{0.08\textwidth}
  	\subfigure[Synthetic Data]{
        \includegraphics[width=0.4\textwidth]{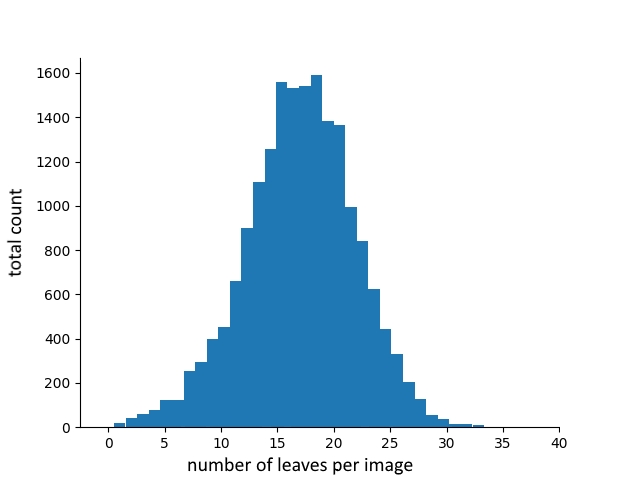}
    }
    \vspace*{2mm}
     \caption{The distribution of the number of leaves per image in the training images. Greater variance is presented in the synthetic data (b) while preserving the mean.}
    \label{fig:leafCounts}
\end{figure}

Synthetic data combined with domain randomization provides the ability to increase the variability across a dataset compared to data augmentation (where the real data is cropped, stretched, blurred, etc). Figure \ref{fig:leafLocDist} shows the distribution of leaf centroid locations between the CVPPP A1 and synthetic data sets. Clearly, in the real dataset (CVPPP A1), the majority of leaves appear around the center of the image. The synthetic data, while biased towards the image center, contains a wider distribution of leaf positions. A wider distribution of data to train on, improves a model's ability to generalize. Figure \ref{fig:leafCounts} shows the distribution of number of leaves per image for CVPPP A1 and synthetic images. While the mean has been matched in the synthetic data, a larger variance and distribution tails demonstrates the wider distribution to learn from.

%
%

\begin{figure}[htpb]
  \centering
    \subfigure[RGB]{
        \includegraphics[width=0.3\textwidth]{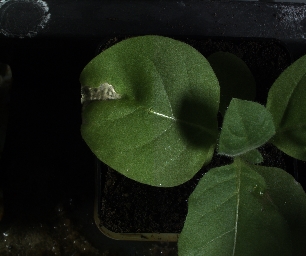}
    }
    \hspace{0.1\textwidth}
  	\subfigure[Prediction]{
        \includegraphics[width=0.3\textwidth]{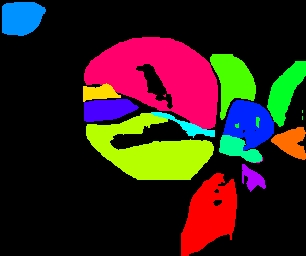}
    }
    \vspace*{2mm}
    \caption{Plant 70 of the A3 test set. Prominent textural edges from the leaf veins lead to over-segmentation of the leaf. Further, invariance to shadowing can be seen as it has no effect on the segmentation.}
    \label{fig:a3plant70}
\end{figure}

Within the models trained on only synthetic data, changing the texture per leaf (\textit{Synthetic-leaf}), performed the best, except in the case of the tobacco plant (A3 in Table \ref{tab:cvpppResults}). The improvement of a per leaf texture over a uniform texture choice per plant led us to consider the basic premise of domain randomization: the model was likely over-fitting to the simplicity and unrealistic distribution of our simulated textures, so increasing the variance in the training data should make the model more invariant to the distribution difference in the test set. However, attempting to add extremely high variance textures (\textit{Synthetic-COCO}), lead to extremely poor results, indicating that plant texture is a important aspect in segmenting leaves. This relevance of texture can also be seen in the comparatively poor performance for the A3 tobacco plant test set. The larger leaves and image resolution for A3 can lead to more visually prominent leaf veins. The over-segmentation of the leaf along a visually prominent vein can be seen in Figure \ref{fig:a3plant70}. Blurring these images before segmentation saw an improvement in performance for these cases, but resulted in the under-segmentation of leaves on the smaller plants. Resilience to the distinct shadow on the largest leaf (Figure \ref{fig:a3plant70}) can also be seen. The lack of over-segmentation is likely due to the synthetic data containing hard shadowing effects from the 3D geometric plant models and the rendering pipeline. 

The higher performance of the \textit{CVPPP-A1} model trained on the tobacco (A3) test set, indicates the real data contains important textural information not present in the simulated data, which leads the model to not over-segment the tobacco leaves. The best results across all CVPPP LSC test sets were from training with mixed batches containing both real and synthetic data. We conclude this is due to the textural information from the real data, and the increased geometric diversity from the simulated plant data. 

In addition we note that training on real and synthetic data also helps reject false positive detection on the background (e.g. Figure \ref{fig:cvpppA1_plant117}). In this case, the synthetic data did not contain any backgrounds with any black and white fiducial markers. The extra background variance across the combined set may have enabled the model to not detect this artifact.

These observations indicate that further performance benefits could be derived by increasing the variation of both the background images and the foreground textures. A simple method to increase background variation would be to physically photograph the nursery before the plants start growing. In addition distractor objects (such as the white fiducial marker) could be added to the foreground \cite{Dwibedi2017}.
The foreground leaf textures used in our experiments were sampled from a small, manually created, set of 30 leaf crops. Increased performance could then be expected by leveraging more sophisticated textures, both increasing the variation and perhaps by making the image more photo-realistic. There are several leaf databases that would be suitable for extracting this texture data \cite{leafsnap_eccv2012}. 

Furthermore, to increase the variance in textures, further investigations will consider more sophisticated approaches synthesizing leaf and plant geometry leveraging L-systems.
The methods used in this paper were all based off a single inspiration leaf of an \textit{Arabidopsis} plant, with relatively compact leaf shapes. Tobacco plants (A3), however, have  broader leaves where are not as well separated from each other. By creating training data sets with larger leaf and plant geometry variance, or at least matching the geometry to new target domains, we expect further increases in performance. This may also include incorporating leaf vein structures in the synthetic data to address the phenomenon presented in Figure~\ref{fig:a3plant70}. 

Our generated synthetic dataset is publicly available at \footnote{\url{https://research.csiro.au/robotics/databases}}. The synthetic dataset contains 10,000 top down images of synthetic \textit{Arabidopsis} plants and their corresponding 2D segmentation labels.

\section{Conclusion}
\label{sec:conclusion}

We have presented and evaluated a framework for generating synthetic \textit{Arabidopsis} images and accompanying instance segmentation labels. By inspiring geometry and sampling real world textures for background and foreground, we have outperformed state-of-the-art results achieving 90\% symmetric best dice score on the CVPPP A1 test set. Furthermore, this framework has the potential to be rapidly adapted to a new plant and automate leaf segmentation within a plant phenotyping setting. We achieved best results across all test sets, except CVPPP A3, when training on both the real data and synthetic data. Presumably, this is due to the extra textural information from the real data and the increased geometric diversity from the simulated plant data.

Future work will be directed at improving the variation in geometry and texture of the synthetic data. We would expect a performance and generalisation increase by sampling leaf textures from a wider, domain specific distributions. For geometric variation, supporting a structure description framework such as L-systems would alleviate the ability to simulate different plant species.\\

\textbf{Acknowledgements:} This research was funded by the CSIRO AgTech Cluster for Robotics and Autonomous Systems. The authors thank the Robotics and Autonomous Systems team for their valuable feedback and discussions, notably Mark Cox, Kamil Grycz, Terry Kung and Micheal Thoreau. Further, we thank Robert Lee for his advice and support.
\clearpage

%
%
\bibliography{egbib}



\section*{Supplementary material}

# Synthetic Arabidopsis Dataset

Daniel Ward and Peyman Moghadam

Robotics and Autonomous Systems, CSIRO Data61, Brisbane, Australia
*firstname.lastname@data61.csiro.au*

**Abstract**

We introduce a synthetic dataset of 10,000 top down images of *Arabidopsis* plants. Leaf instance segmentation labels for each image are also presented. This dataset was designed to accompany the real dataset provided with the Leaf Segmentation Challenge of the Computer Vision Problems in Plant Phenotyping. Furthermore, we release a leaf instance segmentation pretrained model based on the Mask-RCNN architecture. More information about the dataset can be found at: https://research.csiro.au/robotics/databases/synthetic-arabidopsis-dataset/.

## 1 Introduction

Precision agriculture involves using advanced technology, such as imaging and sensing systems, to observe, automate and improve agricultural systems [1]. Plant phenotyping is the identification and evaluation of plant properties and traits under different environmental conditions. It can provide important information in seed production and plant breeding settings and hence, is an important aspect of precision agriculture.

The Computer Vision Problems in Plant Phenotyping (CVPPP) workshop[1] has been held to address the challenges and extend the state of the art in the use of computer vision in plant phenotyping [2]. Further, the Leaf Segmentation Challenge[2] (LSC) of the CVPPP aims to benchmark and advance the state of the art in leaf segmentation, an prominent part of plant phenotyping [3].

We present a synthetically generated dataset of *Arabidopsis* plants. This is designed to accompany data presented with the CVPPP LSC. Our dataset contains top down view renders of synthetically generated plants similar to the real images presented in [2]. We also release a pretrained model for leaf instance segmentation based on the Mask-RCNN architecture [4].

## 2 The Dataset

Within this section, our synthetic dataset is introduced and described. Ward *et al.* [3] outlines the details of the design and production of this dataset. The images and the leaf instance segmentation labels are presented in Section 2.1. A number of example data samples are visualised in Section 2.3.

### 2.1 Data Description

The synthetic *Arabidopsis* dataset contains 10,000 top down images (width×height: 550×550 pixels). All images are stored in PNG format.

*Leaf Instance Segmentation Labels:* Each RGB image has a corresponding leaf instance segmentation annotation. These are the same size as the RGB images and stored in PNG format. Each leaf in an image is uniquely identified by a single colour. A pixel with value 0, depicted as black, corresponds to the background in the image. By converting the segmentation label to a 2D grayscale image and obtaining a set of unique values (excluding 0 as background), the pixel values for each individual leaf can be obtained. To produce a mask for a single leaf, all pixels but those of a value equal to that of the desired leave must be set to 0.

*File types and naming conventions:* This data follows the same naming convention as the CVPPP LSC [2]. The filenames have the form:

---

[1] https://www.plant-phenotyping.org/cvppp2018
[2] https://competitions.codalab.org/competitions/18405

- **plantXXXXX_rgb.png**: the RGB colour image of the synthetic data;
- **plantXXXXX_label.png**: the corresponding leaf instance segmentation label;

where **XXXXX** is an integer number representing the sample number within the dataset.

## 2.2    Obtaining The Data

This dataset can be downloaded from [5]. When making use of this data we ask that [3, 5] are cited.

## 2.3    Example Data Visualisations

Figure 1 displays eight example images and their corresponding leaf instance segmentation label.

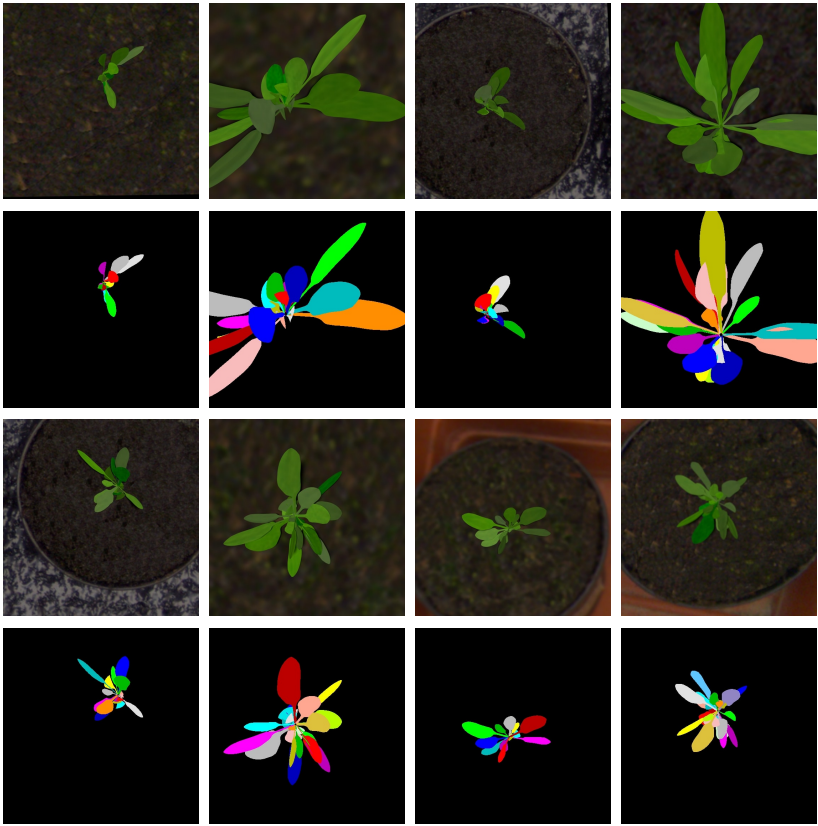

Figure 1: A visualisation of selected images from the synthetic *Arabidopsis* dataset with their corresponding leaf segmentation labels.

# 3 Pre-trained Model and Code

A leaf segmentation model trained on this synthetic data is also available. The deep learning model can be downloaded from [5]. Furthermore, a python script to run this model is available at: `https://bitbucket.csiro.au/scm/ag3d/leaf_segmenter_public.git`. The model requires Matterport's [6] implementation of Mask-RCNN [4]. Instructions for setting up and running the code can be found in the code repository readme file.

# 4 Future Work

Current leaf segmentation pipeline works only on 2D top-down RGB images. In Future work, we develop new segmentation methods that could work on other modalities such as Thermal [7, 8, 9, 10], RGB-D [11, 12, 13] and hyperspectral [1].


\end{document}